\documentclass[letterpaper, 10 pt, conference]{ieeeconf}  

\IEEEoverridecommandlockouts                              

\usepackage{graphics} 
\usepackage{epsfig} 
\usepackage{mathptmx} 
\usepackage{times} 
\usepackage{amsmath} 
\usepackage{amssymb}  
\usepackage{balance}
\usepackage{siunitx}
\usepackage{comment}
\usepackage{xcolor}
\usepackage{dirtytalk}
\usepackage{gensymb}
\usepackage{algorithm}
\usepackage{algpseudocode}
\usepackage{accents}
\usepackage{makecell} 
\usepackage{float}
\usepackage{cite}

\title{\LARGE \bf
A Wearable Resistance Device's Motor Learning Effects in Exercise
}

\author{Eugenio Frias-Miranda$^{1}$,  
        Hong-Anh A. Nguyen$^{2}$,     
        Jeremy Hampton$^{2}$,         
        Trenner A. Jones$^{1}$,          
        Benjamin Spotts$^{1}$, \\     
        Matthew Cochran$^{3}$,        
        Deva D. Chan$^{2}$,           
        and Laura H. Blumenschein$^{1}$
\thanks{$^{1}$School of Mechanical Engineering, Purdue University, West Lafayette, IN 47907, USA}
\thanks{$^{2}$Weldon School of Biomedical Engineering, Purdue University, West Lafayette, IN 47907, USA}
\thanks{$^{3}$School of Medicine, Indiana University, Indianapolis, IN 46202, USA}
\thanks{Email: \tt\footnotesize \{efrias, lhblumen\}@purdue.edu }
}

\begin{document}

\maketitle
\thispagestyle{empty}
\pagestyle{empty}

\begin{abstract}

The integration of technology into exercise regimens has emerged as a strategy to enhance normal human capabilities and return human motor function after injury or illness by enhancing motor learning and retention. Much research has focused on how active devices, whether confined to a lab or made into a wearable format, can apply forces at set times and conditions to optimize the process of learning. However, the focus on active force production often forces devices to either be confined to simple movements or interventions. As such, in this paper, we investigate how passive device behaviors can contribute to the process of motor learning by themselves. Our approach involves using a wearable resistance (WR) device, which is outfitted with elastic bands, to apply a force field that changes in response to a person's movements while performing exercises. We develop a method to measure the produced forces from the device without impeding the function and we characterize the device's force generation abilities. We then present a study assessing the impact of the WR device on motor learning of proper squat form compared to visual or no feedback. Biometrics such as knee and hip angles were used to monitor and assess subject performance. Our findings indicate that the force fields produced while training with the WR device can improve performance in full-body exercises similarly to a more direct visual feedback mechanism, though the improvement is not consistent across all performance metrics.  Through our research, we contribute important insights into the application of passive wearable resistance technology in practical exercise settings.

\end{abstract}
\section{Introduction}

As the field of human movement science has advanced, it has shown promise for enhancing and improving motor skills \cite{shishov2017parameters, zhang2019wearables}. Motor learning is defined as the natural acquisition and refinement of motor skills \cite{tassignon2021exploratory}. To artificially affect motor learning, feedback signals, including visual, auditory, and kinesthetic force cues, have been developed to facilitate skill acquisition and enhance movement precision and efficiency \cite{sigrist2013augmented}. In particular, kinesthetic force feedback is an attractive and effective option for motor learning because it directly interacts with the physical aspects of movement execution and adaptation \cite{culbertson2018haptics}. Traditionally, the study of motor learning through force feedback has relied on large, cumbersome kinesthetic force feedback devices to provide continuously controlled forces across all degrees of freedom of a target exercise \cite{lee2010effects, basalp2021haptic}. Properly designed kinesthetic force feedback devices facilitate motor learning even in the context of artificial restrictions such as visuomotor rotations \cite{losey2019improving}. However, while these lab results hold potential for indicating how motor learning can be affected, their translation to real-world contexts has not yet been achieved.


\begin{figure}[tb]
    \centering
    \includegraphics[width=0.8\columnwidth]
    {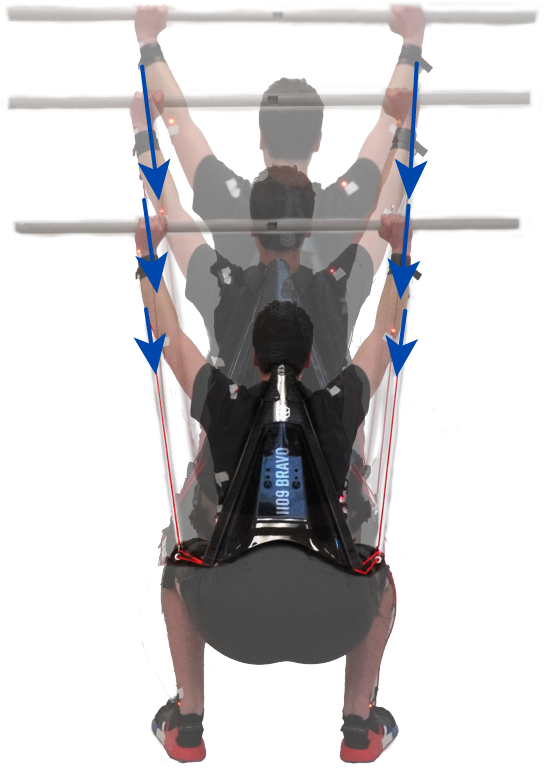}
\caption{An individual performing an overhead squat while wearing a wearable resistance (WR) device. The device applies force feedback through resistance bands. As the user squats down, the resistance force decreases.}
    \label{fig:firstPager}
    \vspace{-1.5em}
\end{figure}

Wearable devices have begun to address this gap by providing force feedback during more realistic everyday motions. While some devices have been designed to actively control all degrees of freedom, most wearable devices choose to target specific joints or pieces of motions \cite{park2023design,chen2021novel,conner2021pilot}. These wearable kinesthetic force feedback devices have demonstrated usefulness in rehabilitation \cite{blank2014current,conner2021pilot, kim2024soft}, in reducing the metabolic rate and muscle fatigue during exercise \cite{chen2021novel}, and in improving personal fitness \cite{park2022design}. These works have focused on optimizing the force profiles for the active degrees of freedom, either to maintain human engagement \cite{blank2014current} or to reduce metabolic costs \cite{chen2021novel}. Still these devices often end up either overly complex or needing to be optimized for each given movement. Contrarily, looking to passive force application, resistive and plyometric exercise alone have been shown to help increase performance in sports such as tennis \cite{centorbi2023resistance} and to increase neuromuscular fitness \cite{novak2023effects}. This suggests that properly designed passive force fields may yield some benefit to motor learning without needing to actuate each degree of freedom. 


In pursuit of this, we examine the effects of the force field produced by a wearable resistance (WR) device on motor learning (Figure~\ref{fig:firstPager}). In this paper we investigate the integration of WR devices with motor learning. In Section~\ref{sec:Force Sensor}, we design and calibrate a stretch sensor to measure the forces generated by the WR device, and show how these forces vary in practical exercise scenarios. Then, in Section~\ref{sec:study} we present the results of a study on the application of the WR device in enhancing the motor learning process for a single target exercise, overhead squats. The concluding discussions reflect on the potential implications of these findings for both the field of motor learning and the practical deployment of WR devices in exercise.

\section{Force Field of a Wearable Resistance Device}
\label{sec:Force Sensor}
\subsection{A Stretch Sensor for Resistance Bands}

\begin{figure}[tb]
    \centering
    \includegraphics[width=1.0\columnwidth]{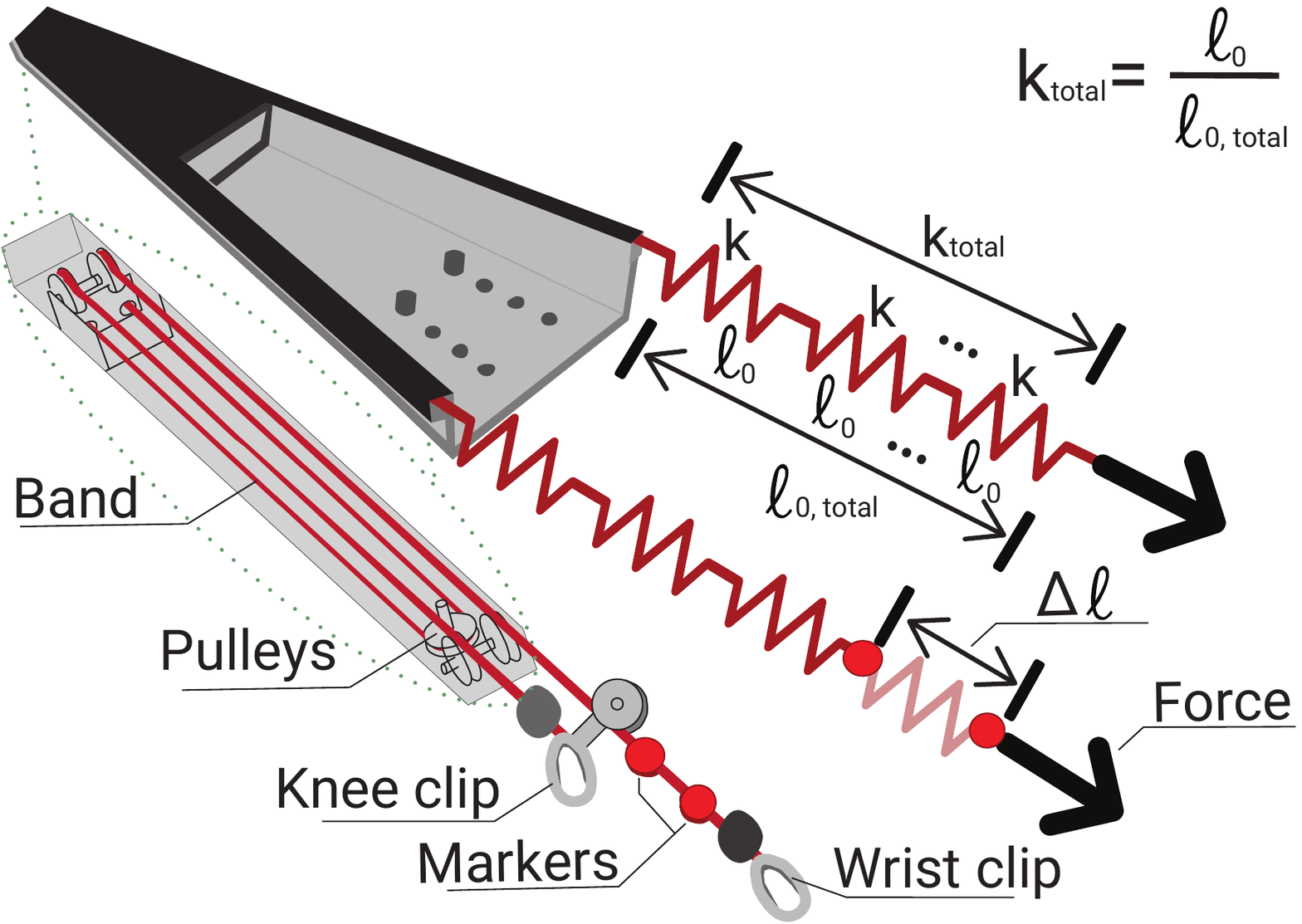}
\caption{Illustration of the wearable resistance (WR) device: (lower left) a cross-sectional view of one of the sides showcasing the resistance band wrapping around the pulleys and the knee/wrist clips, (right) an isometric view of the WR device is shown with the resistance bands presented as a series of springs showing the initial length of the location of the sewing of the markers $\ell_{0}$, the difference between the length of the motion capture markers throughout the exercises and length of the motion capture markers at rest $\Delta \ell$, and the average between both of the band's calibration stiffness $k_{cal}$.}
    \label{fig:WRDeviceOverview}
    \vspace{-0.5em}
\end{figure}

For this study, we used a specific WR device, the NeuroPak (1109 Bravo, KY, USA), designed to be used in a wide range of athletic activities (Figure~\ref{fig:WRDeviceOverview}). The WR device is worn like a backpack and functions through a series of pulleys and elastic resistance bands (SGT KNOTS Supply Co., NC, USA) that connect from the backpack-like frame to straps placed around both the knees and wrists using clips. This setup with clips from the frame to the knees preloads the elastic bands to maintain a constant force presence. 

To measure the force applied by the WR device to the user's wrists, motion capture markers (Impulse X2E System. PhaseSpace, San Leandro, CA) were sewn to the WR device's bands to measure the band stretch. Considering the WR device's bands as a series of interconnected springs, we can use Hooke's law and, with the experimental displacement and band stiffness, obtain the resistance force applied, $F$. We measure the experimental displacement via these motion capture markers and band stiffness determined in the calibration experiment discussed in the next section. This stiffness-displacement relationship was captured in the following modified stiffness equation:
\begin{equation}
F = k_{cal} \cdot \frac{\ell_{cal}}{\ell_{0}}  \cdot \left( \Delta \ell \right) + F_{i}
\label{eqn:stiffness}
\end{equation}
where $k_{cal}$ is the band's calibration stiffness, $\ell_{cal}$ is the calibration experiment segment length, $\ell_{0}$ is the segment length between the motion capture markers with zero force applied, $\Delta \ell$ the relative stretch of the segment length throughout the exercises as measured by the motion capture markers, and $F_{i}$ is the band's initial force observed in the calibration experiments. The same WR device with the same resistance bands was used in the calibration experiments and throughout the entirety of this paper to maintain consistency.

\subsection{Calibration of the Stretch Sensor}

\begin{figure}[tb]
    \centering
    \includegraphics[width=1.0\columnwidth]{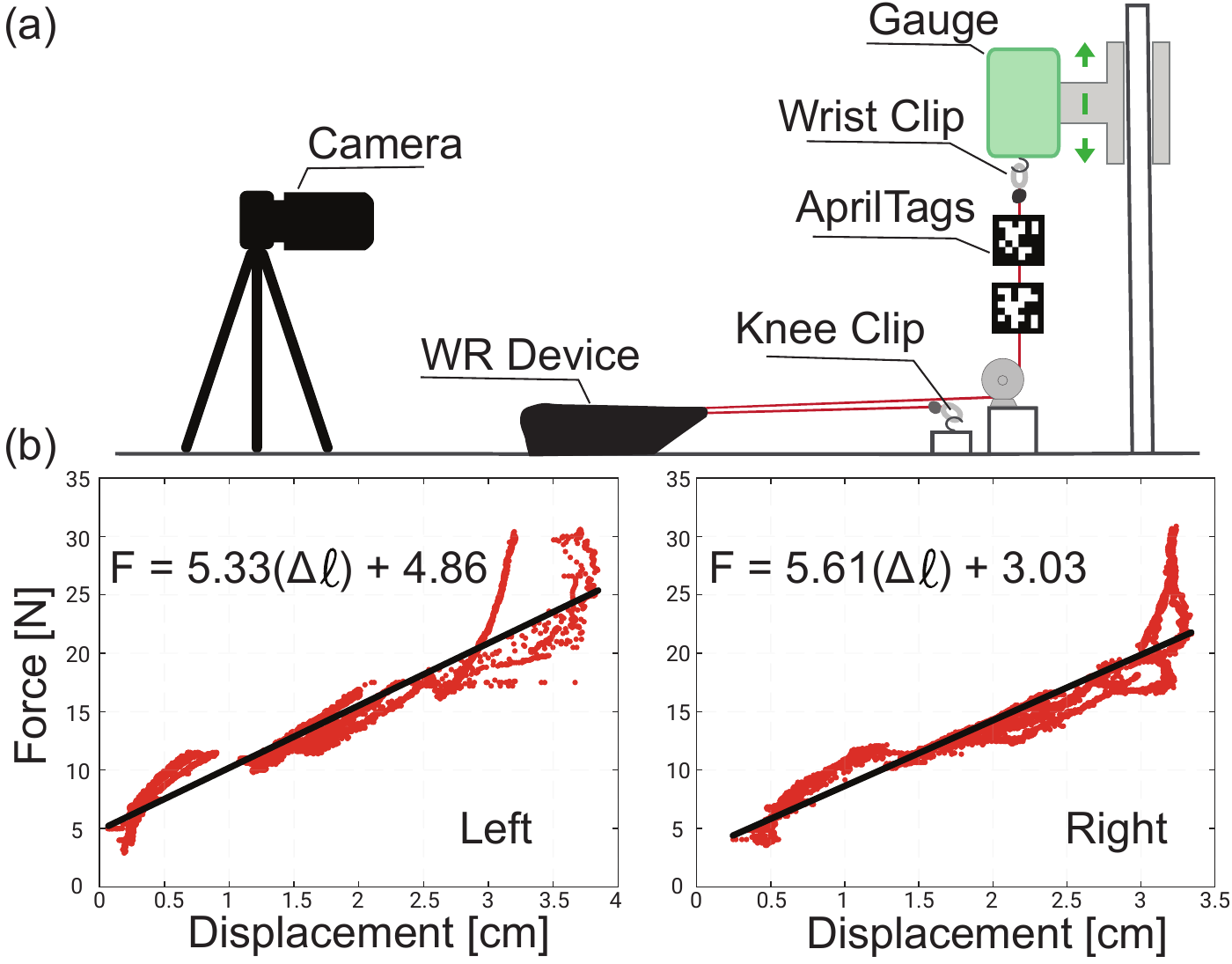}
\caption{The calibration experiments experimental setup (a) and calibration results of the stretch sensor (b). (a) The force gauge (Series 7, Mark-10 Corporation, NY, USA), camera, AprilTag markers placement, wrist clip, and knee clip. (b) The force-displacement relationship for the stretch sensor, with separate lines representing the left $(F = 5.33*\Delta \ell + 4.86[N])$ and right $(F = 5.61*\Delta \ell + 3.03[N])$ sensors. The x-axis is displacement in cm, and the y-axis is force in N.}
    \label{fig:calibrationStretchSensorResults}
    \vspace{-1.5em}
\end{figure}

\begin{figure*}[tb]
    \centering
    \includegraphics[width=1.8\columnwidth]{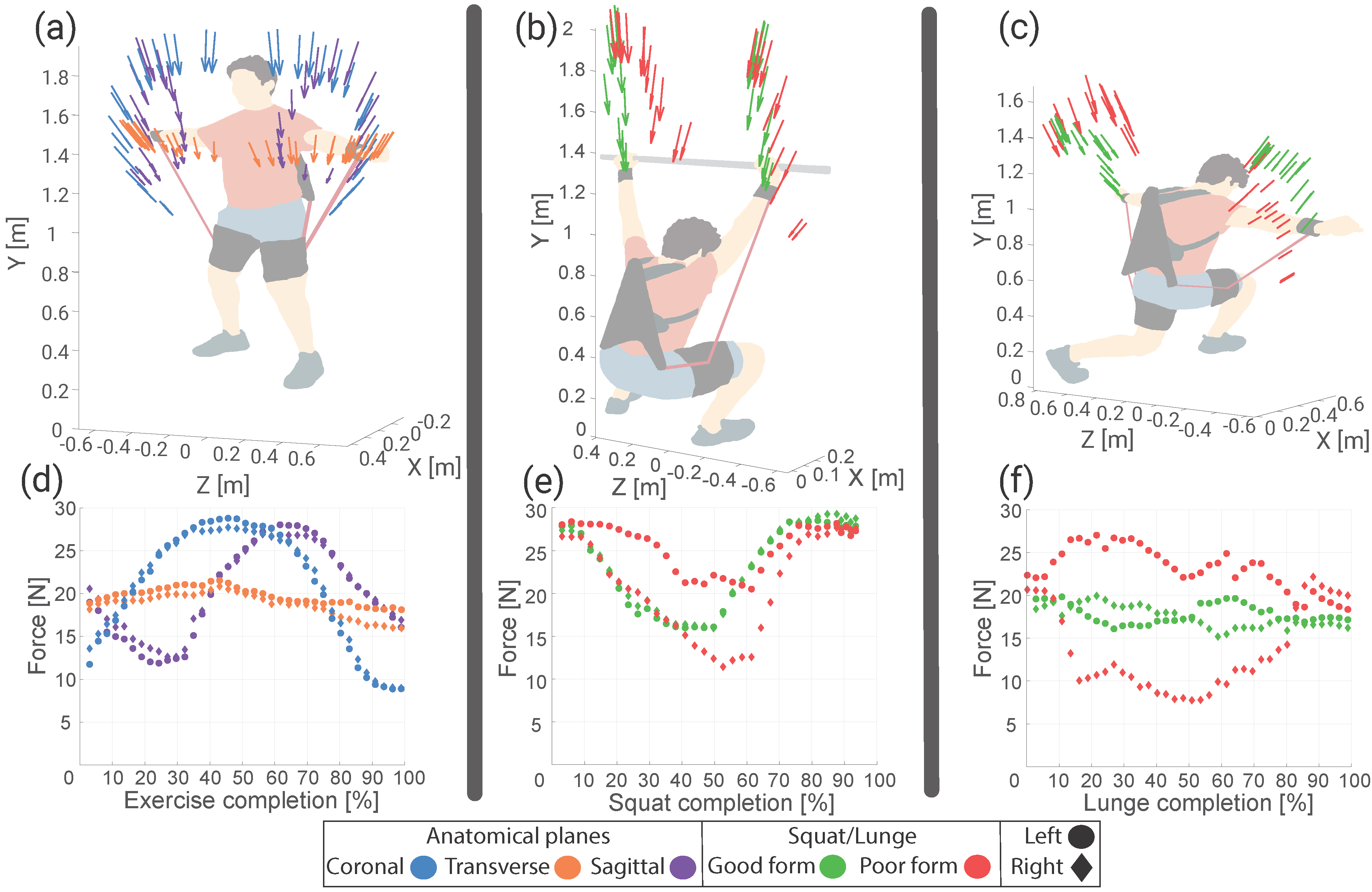}
\caption{(a) force vector plot of wearable resistance (WR) device while performing anatomical plane exercises (coronal, transverse, and sagittal) (b) force vector plot of WR device while performing overhead squats with good and poor form (c) force vector plot of WR device while performing extended arm lunges, with good and poor form (d) resistance band force over exercise completion percentage for anatomical plane exercises (e) resistance band force over squat completion percentage (f) resistance band force over lunge completion percentage }
    \label{fig:forceDemo}
    \vspace{-1.5em}
\end{figure*}

A set of experiments were developed to obtain the resistance band stiffness. A force gauge (Series 7, Mark-10 Corporation, NY, USA) was employed to test the WR device's band at different force ranges. Using the setup shown in Figure~\ref{fig:calibrationStretchSensorResults}(a), each resistance band (right and left side) was pulled in 4 overlapping force intervals to allow for a complete recording of the band's force-displacement behavior. Before data collection, the resistance bands were pretensioned with one of three tensile loading cycles at each of the 4 intervals. AprilTag markers provided exact measurements of the bands' elongation under different loading conditions. Given that the band materials are expected to be hyperelastic, the boundaries of the experiment were expanded until the s-curve relationship emerged between the band's displacement and the force. The force-displacement data from the calibration experiment can be seen in Figure~\ref{fig:calibrationStretchSensorResults}(b). 

Due to the nature of this experiment, Equation~\ref{eqn:stiffness} is set-up to accommodate different lengths ($\ell_{0}$) of the band while maintaining an accurate band resistance stiffness. Additionally, we add an initial force to the stiffness equation, which accounts for the initial hyperelastic behavior of the bands. Since there is not significant variation expected between bands, the right and left calibration variables, ($\ell_{cal}$, $F_{i}$, and $K_{cal}$) were averaged and used for both sides. 



\subsection{Force Field Demonstration}

With a developed stretch sensor and calibration methodology for the WR device, we established a series of exercises aimed at evaluating the device's functionality. This included a selection of two widely practiced workout exercises, modified to incorporate extended arm positions, thereby optimizing the impact of the WR device. This routine was composed of (1) quasi-static arm exercises, which were conducted in the coronal, sagittal, and transverse anatomical planes, and (2) dynamic exercises, specifically overhead squats, and extended arm lunges. The quasi-static exercises provided a controlled environment to assess the range of forces exerted by the WR device, while the dynamic exercises allowed us to observe the device's performance in real-life, movement-based scenarios.

For the dynamic exercises, we paid particular attention to the comparison between good and poor forms, especially noting the impact of pelvic tilt during squats and lunges. Our analysis of the force vector plots revealed several key observations, Figure~\ref{fig:forceDemo}a-Figure~\ref{fig:forceDemo}c. Firstly, all force vectors were directed inward towards the user, corresponding to the direction of the resistance bands. We also noticed that the magnitude of these force vectors increased as the user raised their arms, while a decrease was observed whenever there was knee flexion in the dynamic exercises. Consistent forces were recorded between the left and right bands, except in instances where the dynamic exercises were performed with poor form.

The results from the individual exercises provided further insights into the characteristics of the force field generated by the WR device, Figure~\ref{fig:forceDemo} d-Figure~\ref{fig:forceDemo}f. The coronal plane exercise demonstrated a negative sinusoidal force pattern, while the sagittal plane exercise displayed a bell-shaped force curve. The transverse plane exercise showed a constant force of around 20 N, consistent with the hands being maintained at chest level throughout the exercise.

The squat exercise, when performed with good form, exhibited a negative bell-shaped force curve. However, when performed with poor form, the force patterns differed significantly between the left and right bands; the left-hand band displayed a positive bell-shaped force curve, and the right-hand band showed a negative bell-shaped curve. This difference in force application suggests the WR device's potential to assist with the correction of asymmetries in movement.

Similar observations were made in the lunge exercise. With good form, the force remained constant at about 18 N, attributed to the steady hand position near chest level throughout the exercise. In contrast, the lunge with poor form presented a disparity in force application, much like the squat with poor form, with the left-hand band force showing a positive bell-shaped curve and the right-hand band a negative one.


\subsection{Discussion}

While wearing the WR device, the dynamic exercises performed with poor form showed an increase in force on the side with increased pelvic obliquity angle, highlighting the device's capability to impose additional resistance in response to incorrect postures. This behavior mimics a force fields that augments errors, thereby encouraging corrective action \cite{israely2016error}. However, there is no perfect direct correlation with task error, the forces in the WR device are also position-centric, this can be seen in the transverse anatomical plane exercise, where the force is constant.

With regards to the other exercises, we can see that the force is significantly influenced by the location of the hands in the full task space. Squatting shows similar forces to coronal arm movements and lunging shows similar forces to the transverse arm movements. This indicates that for normally performed exercises, the forces at the hands are similar to static force fields,exerting a position-dependent force, irrespective of user motion or trajectory. Prior research has also demonstrated that such force fields can improve proprioception, particularly in hand motion training \cite{goble2010plastic}.


The ability of this WR device to both challenge and guide users through physical resistance is similar to how traditional motor learning studies define antagonistic and static force fields. This strengthens our assertion of the WR device's capacity to provide passive force fields could offer the same benefits, as seen in motor learning studies that use antagonistic and static force fields, while also influencing a broader range of motion.

\section{Motor Learning Study on Squatting}
\label{sec:study}
\subsection{Methods}

\begin{figure}[tb]
    \centering
    \includegraphics[width=1.0\columnwidth]
    {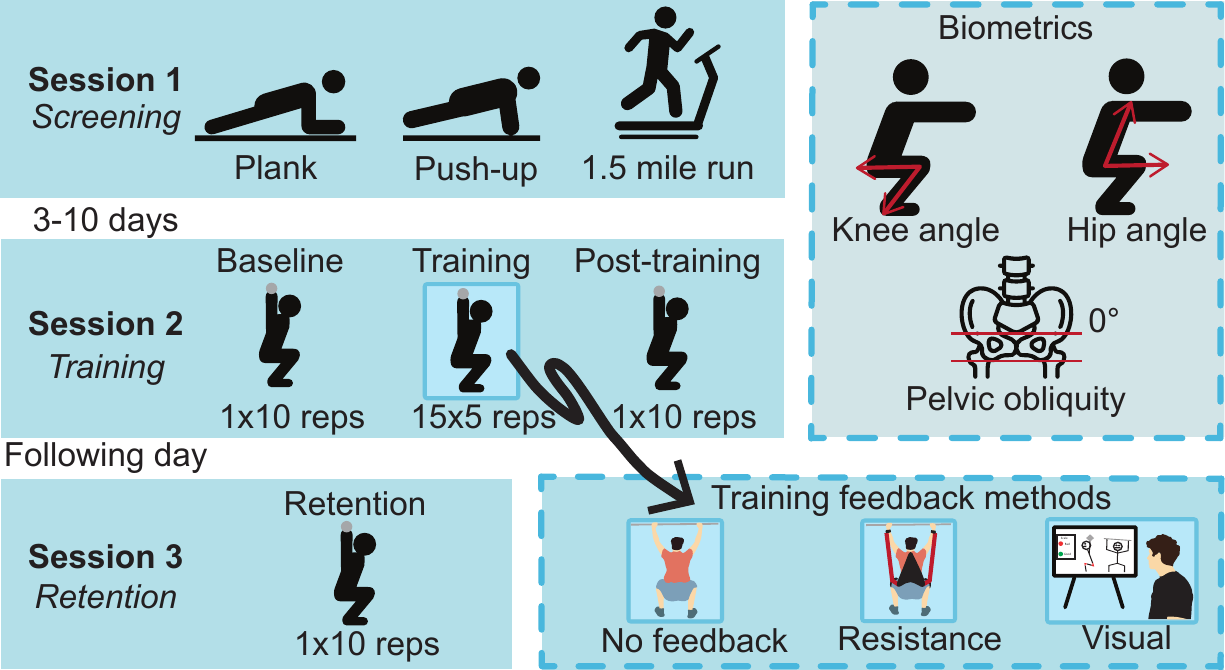}
\caption{Diagram of the study's timeline which includes: session 1 the screening, session 2 training, and session 3 retention. Session 2 involves performing a series of overhead squats: 1 set of 10 reps (baseline), 15 sets of 5 reps (training), and 1 set of 10 reps (post-training). Feedback methods for the training section are no feedback, resistance feedback, and visual feedback. The evaluated biometrics throughout the study are defined: pelvic obliquity, and knee/hip angles. }
    \label{fig:experimentFlow}
    \vspace{-1.5em}
\end{figure}

This study was designed to provide a comprehensive evaluation of the impact of the WR device (NeuroPak, 1109 Bravo, KY, USA) by comparing it to different feedback mechanisms and evaluating them for motor learning, specifically focusing on overhead squats performed by generally athletic individuals. Approved by Purdue University's Institutional Review Board (IRB $\#$2022-1720), 36 subjects completed the entirety of this study, each voluntarily participating and fully briefed on the study's objectives and potential risks. Our experimental design was structured across three sessions (Figure~\ref{fig:experimentFlow}, cumulatively spanning approximately 3.25 hours. Our primary goal was to assess and compare the efficacy of three feedback methods: no feedback, visual feedback, and resistance feedback. 

\begin{figure}[tb]
    \centering
    \includegraphics[width=1.0\columnwidth]{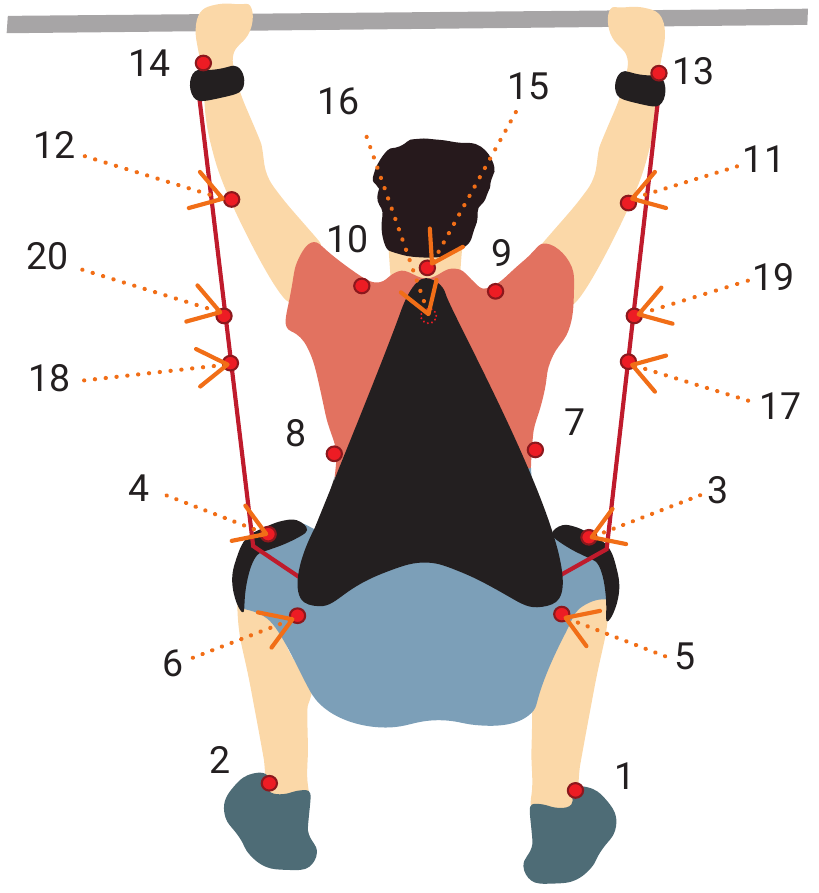}
\caption{Illustration of an individual using a wearable resistance (WR) device while performing an overhead squat. Motion capture markers are labeled in corresponding locations enumerated 1 through 20. Markers 17 through 20 are sewn on the WR device's resistance bands.}
    \label{fig:methods}
    \vspace{-1.5em}
\end{figure}

The initial session functioned as a screening test. Here, subjects were required to provide basic demographic information and undergo a series of athletic screening tests based on the U.S. Air Force's physical fitness test (1.5-mile run, 2 minutes of hand-release push-ups, and a timed plank) \cite{USAFfitnessProgram} to evaluate each participant's physical fitness and suitability for the study(Table~\ref{tab:athletic_test_info}). In the second session, those who got at least 90\% of a passing score the initial screening were allocated to one of the three experimental groups separated by feedback modality. Subjects then performed overhead squats while tracked by 20 motion-tracking markers (Table~\ref{tab:phasespace_led_markers} and Figure~\ref{fig:methods}). This session began with a baseline segment (1 set of 10 squats), followed by the training segment (15 sets of 5 squats) under the assigned feedback intervention, and concluded with a post-training segment (1 set of 10 squats). Subjects were instructed via a video to perform deep squats (knee angle of 120 degrees) and to maintain good form by maintaining an upright posture and symmetric form while doing so. The following day, subjects reconvened for the third session, where they performed additional squats to test 24-hour retention (1 set of 10 squats).

The feedback conditions varied solely during the training segment portion of the study. The resistance feedback group wore the WR device, which provided continuous forces during the squat. The visual feedback group received feedback via a graphical user interface (GUI) at the end of each set of 5 squats during training. The GUI provided binomial colored feedback (green representing good form and red representing poor form) for each of the three measures of form in the study. Visual feedback represents a good comparison as prior work has shown evidence for visual GUI feedback improving squatting form and motor learning \cite{bonnette2020technical, kal2023optimal}. All biometric data was collected using motion capture, and an exit survey was administered to capture subjects' subjective feedback on their training experience.

\begin{table}[h]
    \centering
    \caption{Demographics and Results of Athletic Fitness Test Qualifying Subjects}
    \label{tab:athletic_test_info}
    \begin{tabular}{lcccc}
    \hline
    & \textbf{All} & \textbf{No feedback} & \textbf{Resistance} & \textbf{Visual} \\
    \hline
    \textbf{Subject [\#]} & 36 & 12 & 12 & 12 \\
    \textbf{Age [yrs]} & 22.6 $\pm$ 4.0 & 23.3 $\pm$ 4.5 & 21.9 $\pm$ 3.2 & 22.8 $\pm$ 4.1 \\
    \textbf{Male [\#]} & 28 & 11 & 8 & 9 \\
    \textbf{Female [\#]} & 8 & 1 & 4 & 3 \\
    \textbf{90-100\%} & 5 (all male) & 2 & 2 & 1 \\
    \textbf{$\geq$100\%} & 31 & 10 & 10 & 11 \\
    \hline
    \end{tabular}
\end{table}

\begin{table}[h]
    \centering
    \caption{Marker locations}
    \label{tab:phasespace_led_markers}
    \begin{tabular}{cc}
    \hline
    \thead{\textbf{Marker} \\ \textbf{[Left $|$ Right]}} & \textbf{Location} \\
    \hline
    2  $|$  1& Ankle (top of lateral malleolus) \\
    4  $|$ 3& Lateral femoral epicondyle \\
    6  $|$ 5& Hip (iliac crest's most lateral aspect) \\
    8  $|$ 7& Side, mid axillary line, at R8 \\
    10 $|$ 9& Shoulder side (3 fingers distal from acromion) \\
    12 $|$ 11& Elbow \\
    14 $|$ 13& Wrist (ulna carpal joint) \\
    15& C4 spinous process \\
    16& Sternum (sternal angle rib 3) \\    
    18 $|$ 17& Sewn to band, close to knee \\
    20 $|$ 19& Sewn to band, close to wrist \\
    \hline
    \end{tabular}
\end{table}

\subsection{Objective Measures}

The motion capture system measured kinematic data during the squat exercises. The study focused on several key biometrics, including knee angle, hip angle, and pelvic obliquity. These metrics were selected due to their relevance in assessing squat form. The three biometrics were approximated using the data from the motion capture markers as follows: knee angle, angle formed between ankles, knees, and hips markers; hip angle, angle between the chest, hip, and knee markers; and pelvic obliquity, angle between the two hips and the flat plane, referring to the angle/misalignment of the pelvis. From these measures, we also calculated and evaluated asymmetries between the left and right side for knee angle difference, and hip angle difference. These parameters assessed the quality of the squat, as they reflect the participant's ability to maintain balance, coordination, and proper body alignment throughout the movement.

\subsection{Hypothesis}

For this study, we had two hypotheses: 1) the WR feedback will improve learning of proper squat form compared to no feedback and on par with visual feedback, and 2) the learning produced by the WR device will be retained after the feedback is removed.  


\begin{figure}[tb]
    \centering
    \includegraphics[width=1.0\columnwidth]
    {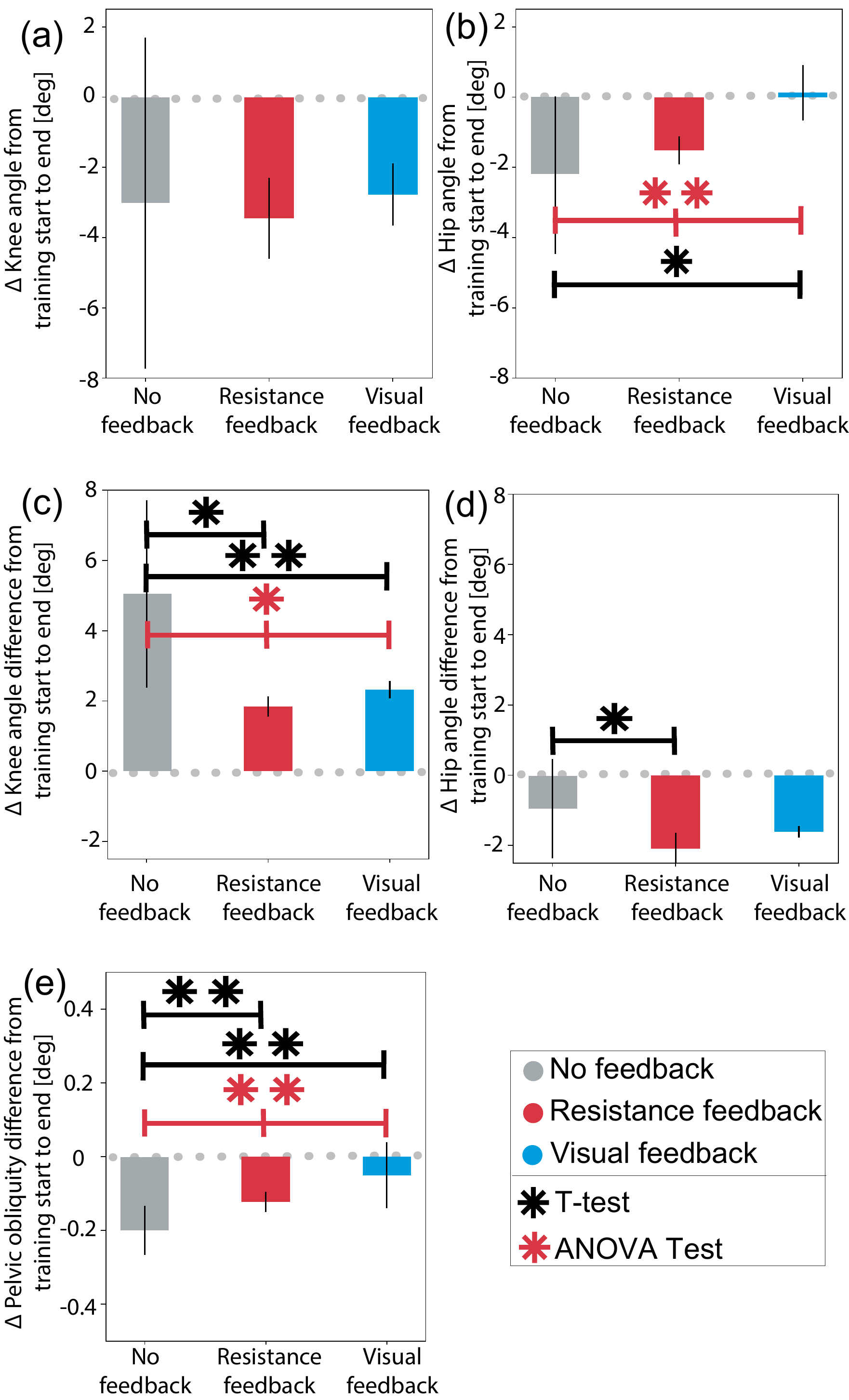}
\caption{Change in measurements from training start to end. Black asterisks with black lines indicate significant differences found with T test, and red asterisks and red lines indicate significant differences found with ANOVA. The notation for T test and ANOVA is as follows: *** for $p < 0.001$, ** for $p < 0.01$, * for $p < 0.05$, and n.s. for not significant.}
    \label{fig:StartEndAsym}
    \vspace{-1.5em}
\end{figure}

\subsection{Statistical Results}
The performance metrics evaluated for all statistical tests were knee angle, hip angle, pelvic obliquity, knee angle difference, and hip angle difference. Statistical significance for all tests described below was defined at $\alpha = 0.05$. Data sets were tested for normality of distribution and equality of variances using the Kolmogorov-Smirnov test and Levene’s test. In cases where data were not normally distributed or had unequal variances, non-parametric tests were used for specific hypothesis testing.

Resistance and visual feedback effects on specified biometrics were evaluated using independent t-tests when data were parametric and Mann-Whitney U test when data were nonparametric. To fully evaluate the effects of feedback methods, comparisons were tested for no feedback vs. resistance feedback, no feedback vs. visual feedback, and visual feedback vs. resistance feedback. Additional ANOVA tests were run to compare no feedback vs. visual feedback vs. resistance feedback data sets for changes from baseline to training start, baseline to post-training, baseline to retention, and training start to training end, averaging the performance over equal sets of 10 squats for each section.

\begin{figure}[tb]
    \centering
    \includegraphics[width=1.0\columnwidth]
    {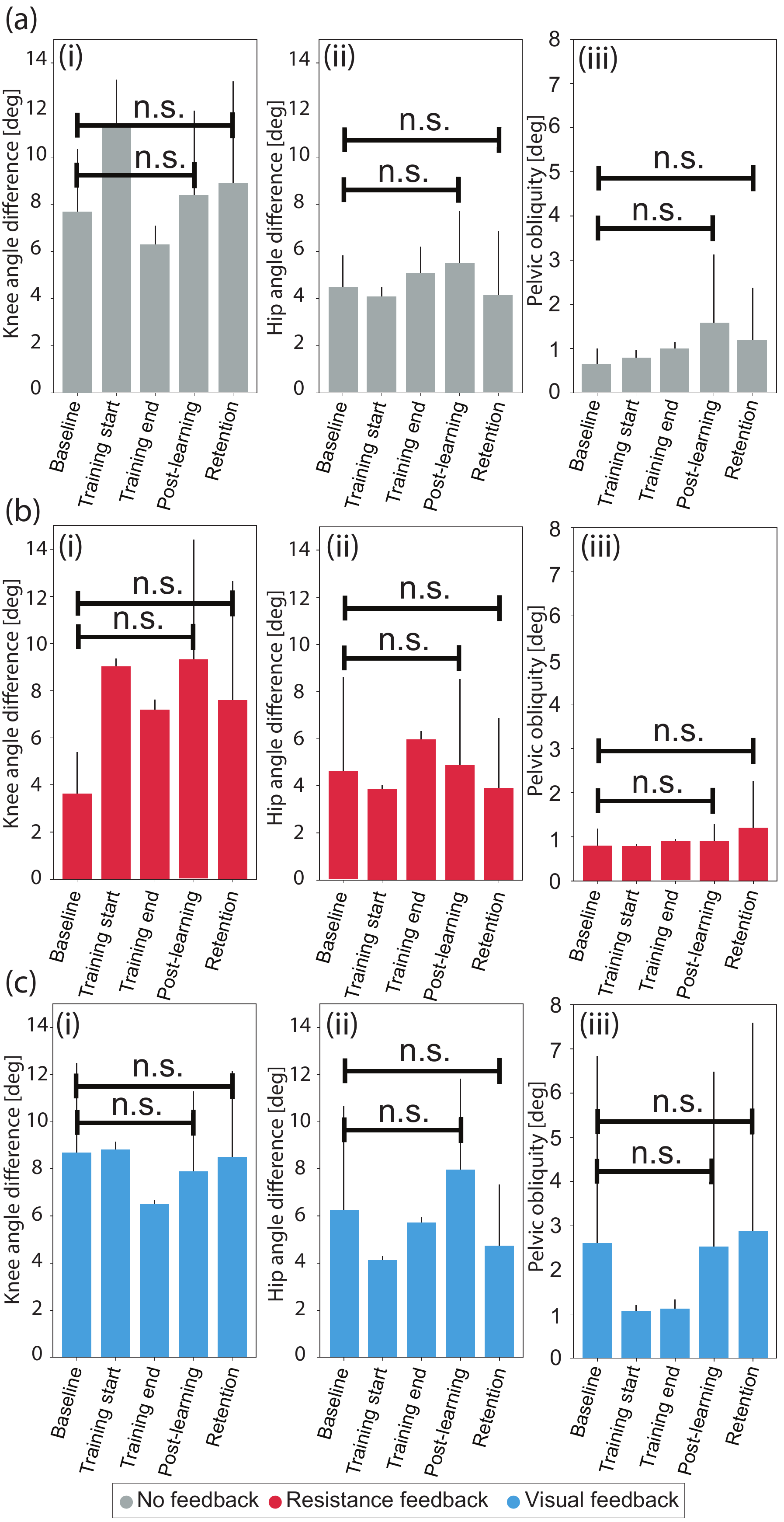}
\caption{Range of motion measurements from baseline to retention. Black asterisks with black lines indicate significant differences found with T-test, and red asterisks and red lines indicate significant differences found with ANOVA. The notation for T-test and ANOVA is as follows: *** for $p < 0.001$, ** for $p < 0.01$, * for $p < 0.05$, and n.s. for not significant.}
    \label{fig:BarPlotsBaseRet}
    \vspace{-1.5em}
\end{figure}

To examine the effects of visual and resistance feedback on the subjects’ form while squatting, the knee and hip angles were evaluated for range of motion and the knee and hip angle differences from left to right and the pelvic obliquities were evaluated for asymmetry. From the training start to training end, significant differences in the change in knee angle difference ($p = 0.03$), hip angle difference ($p = 0.044$), and pelvic obliquity ($p = 0.004$) were found between no feedback and resistance feedback (Figure ~\ref{fig:StartEndAsym}). Comparing the no feedback and visual feedback, we found similar significant differences in the pelvic obliquity ($p = 0.001$) and knee angle difference ($p = 0.015$) changes between training start and training end as well as a significant difference in the $\Delta$ hip angle ($p = 0.012$) (Figure ~\ref{fig:StartEndAsym}b). To evaluate the difference in performance between all three types of feedback, a Kruskal-Wallis ANOVA test was performed with the assumption of equal distributions satisfied by testing each metric using a Kolmogorov-Smirnoff test. With this test, we found there was a smaller change in pelvic obliquity ($p = 0.001$) and hip angle ($p = 0.002$) in the no feedback and resistance feedback compared to the visual feedback and more consistency from training start to end in the knee angle differences ($p = 0.01$) in resistance feedback compared to visual feedback and no feedback (Figure ~\ref{fig:StartEndAsym}). While these measures of asymmetry were significantly different between the training start and end, there were no significant differences found in measures for range of motion (Figure ~\ref{fig:BarPlotsBaseRet}) and no significant differences were found in any measures from baseline to post-learning and baseline to retention. 

\begin{figure*}[tb]
    \centering
    \includegraphics[width=2.0\columnwidth]{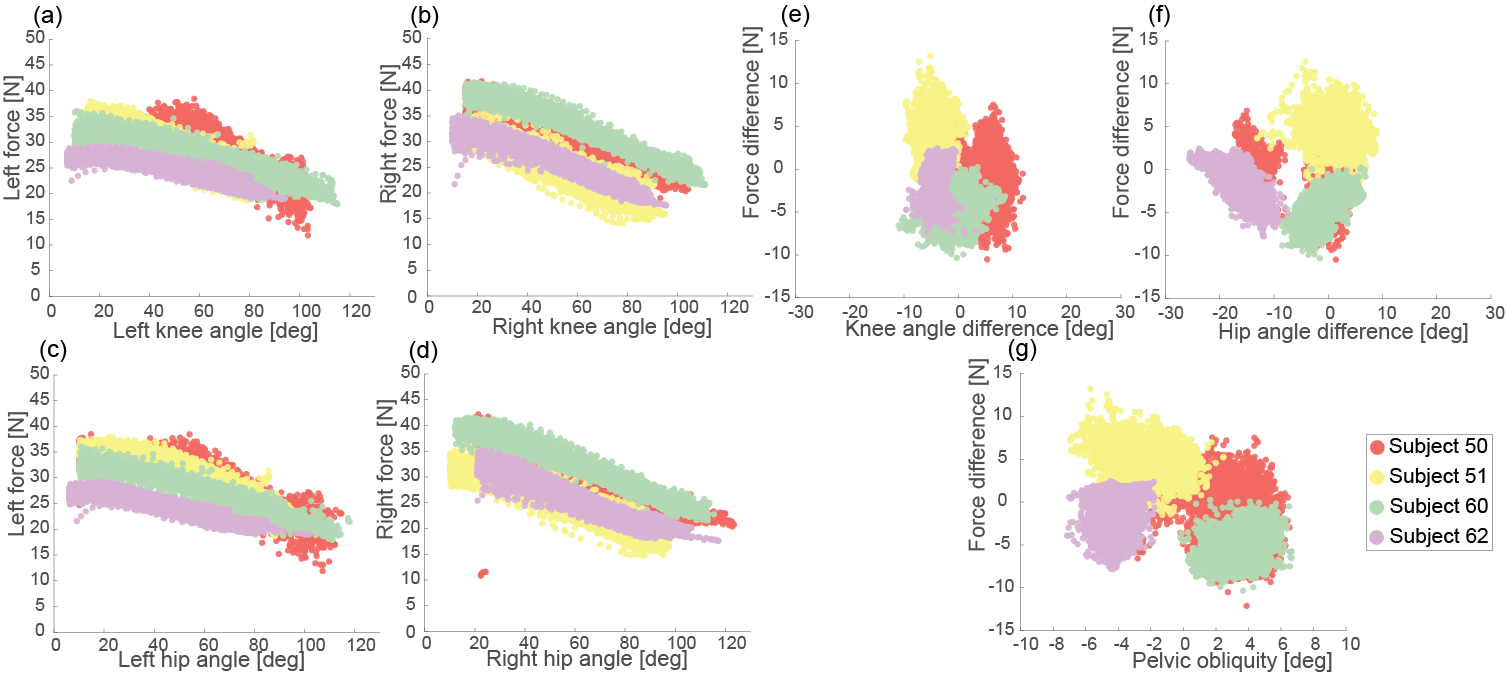}
\caption{Force applied to the subject during squat training as a function of knee angle and hip angle. (a)-(b) Left and right knee angle versus force respectively. (c)-(d) Left and right hip angle versus force. Difference in the right and left forces applied to the subject during squat training over (e) knee angle asymmetry, (f) hip angle asymmetry, and (g) pelvic obliquity.}
    \label{fig:forceAnalysis}
    \vspace{-1.5em}
\end{figure*}

\subsection{Force Analysis Results}

In addition to the objective measures described above, we measured the force applied to participants during the experiments. Due to issues with motion capture markers and occlusions, only 4 of the WR subjects have sufficient force data to analyze during the training segment (Figure~\ref{fig:forceAnalysis}). The force range (between 10-40 N) was perceptible by the subjects based on the post study survey, providing kinesthetic feedback. The data revealed a clear pattern matching with the results in Section~\ref{sec:Force Sensor}: the force applied was lowest when the knee and hip angles were approximately 90-120 degrees, corresponding to the deepest squatting position. As the angles widened towards 0-20 degrees, indicating a return to a standing position, the force increased.

We also examined the force differences (between left and right sides) in relation to asymmetries such as knee and hip angle differences, and pelvic obliquity (Figure~\ref{fig:forceAnalysis}). Our findings indicated that knee angle difference did not influence the force difference during squatting. Similarly, the relationship between pelvic obliquity and force difference was not pronounced, which is not aligned with the differences seen during the ``bad squat" in Fig.~\ref{fig:forceDemo}e. This indicates that the force changes for squat asymmetries may be more complex than the three measures calculated.


\subsection{Discussion}

The statistical results and force analysis of the experiment show that the WR device produces changes in squat performance when compared to no feedback, and that these changes are aligned with the improvements produced by visual feedback. This supports our first hypothesis, that passive WR devices can produce learning on par with active methods. However, these changes are not present for all measures. In particular knee and hip angle for the WR condition was more similar to the no feedback case, while statistically significant improvements were seen in all asymmetry measures for WR feedback compared to no feedback. This may indicate that the force differences experienced while changing squat height, which needed to be remembered between squats, were not as noticeable as force differences generated across the body due to asymmetries. Alternatively, the performance on the knee and hip angle may have started sufficiently high that there was little room for subjects to improve. This conclusion is further supported by the subject survey comments where one subject stated ``I think the resistance feedback helped make bad form harder to do. It also made squats without feedback much easier as there was no resistance" and another commented that the WR device was ``Good for strength training that is not evident/noticeable when working out". Further testing with subjects with a range of athletic abilities may be able to clarify whether training improvements are possible only on asymmetric measures or on all exercise measures. Regardless, these results show that pieces of the WR passive force field can stimulate motor learning changes, so a passive force field designed for an exercise may yield benefits for a passive or active wearable device.

The second hypothesis is not supported by the statistical or force results. There are no significant results seen in the behaviors before and after the training, despite seeing the differences emerge for WR and visual feedback during training. Potentially this indicates that the washout period (i.e. amount of time to unlearn a learned behavior) for the learned improvements was too short to notice when averaging ten squats together.

\section{Conclusion and Future Work}

In this work, we have demonstrated that a passive wearable resistance device can produce force fields with analogs to those seen in motor learning research. These force fields can also induce changes in performance of full-body exercises like squats when compared to similar numbers of squats performed with no feedback. Additionally, this improvement is on par with changes produced by visual feedback. These results indicate that passive force fields may be a beneficial alternative or addition to active feedback devices for motor learning. However, the improvements in squatting were not retained after training and were not consistent across all metrics, indicating that more thought is needed on how to design passive force fields to yield longer results.

In the future, we plan to test these results against other exercise conditions by varying the target exercise and by testing in less athletic populations. Additionally, we will investigate how this force field can be tailored to an exercise or to an individual subject and how the results can support active force feedback devices.

\section{Acknowledgements}

The authors thank the Purdue Ismail Center Staff, Jordan Lyles, and Aaron Farha for help throughout the study.


\bibliographystyle{IEEEtran}
\bibliography{myReferences}

\end{document}